%% file: ms.tex
\PassOptionsToPackage{sort&compress}{natbib}
\PassOptionsToPackage{noend}{algorithmic}
\documentclass{article}

\usepackage{multirow}
\usepackage{pdflscape}
\usepackage{subcaption}
\usepackage{mwe}
\usepackage{tabularx}
\usepackage{rotating}

\usepackage[accepted]{icml2020}

\usepackage{our_style}

\newtheorem{theorem}{Theorem}[section]

\newtheorem{lemma}[theorem]{Lemma}
\newtheorem{definition}{Definition}[section]

\icmltitlerunning{Neuro-Symbolic Visual Reasoning: Disentangling ``Visual'' from ``Reasoning''}

\begin{document}
\include{macros}

\twocolumn[
\icmltitle{Neuro-Symbolic Visual Reasoning: Disentangling ``Visual'' from ``Reasoning''}

\icmlsetsymbol{equal}{*}

\begin{icmlauthorlist}
\icmlauthor{Saeed Amizadeh}{ms}
\icmlauthor{Hamid Palangi}{equal,msr}
\icmlauthor{Oleksandr Polozov}{equal,msr}
\icmlauthor{Yichen Huang}{msr}
\icmlauthor{Kazuhito Koishida}{ms}
\end{icmlauthorlist}

\icmlaffiliation{ms}{Microsoft Applied Sciences Group (ASG), Redmond WA, USA}
\icmlaffiliation{msr}{Microsoft Research AI, Redmond WA, USA}

\icmlcorrespondingauthor{Saeed Amizadeh}{saamizad@microsoft.com}

\icmlkeywords{Visual Reasoning, Visual Question Answering, Differentiable First-Order Logic, Vision-Perception Disentanglement, GQA}

\vskip 0.3in
]

\printAffiliationsAndNotice{\icmlEqualContribution}  

\input{sections/abstract}

\input{sections/introduction}

\input{sections/related}

\input{sections/methodology}
\input{sections/framework}
\input{sections/calibration}
\input{sections/experiments}
\input{sections/conclusion}

\section*{Acknowledgement}
We would like to thank Pengchuan Zhang for insightful discussions and Drew Hudson for helpful input during her visit at Microsoft Research. 
We also thank anonymous reviewers for their invaluable feedback.

\bibliography{ref}
\bibliographystyle{icml2020}

\appendix
\input{sections/Appendix}

\end{document}

%% file: macros.tex
\newcommand{\fwname}{{$\nabla$-FOL}\xspace}

%% file: sections/abstract.tex
\begin{abstract}

Visual reasoning tasks such as visual question answering (VQA) require an interplay of visual perception with reasoning
about the question semantics grounded in perception.
However, recent advances in this area are still primarily driven by perception improvements (e.g.
scene graph generation) rather than reasoning.
\emph{Neuro-symbolic models} such as Neural Module Networks bring the benefits of compositional reasoning to VQA, but
they are still entangled with visual representation learning, and thus neural reasoning is hard to improve and assess on
its own.
To address this, we propose (1) a framework to isolate and evaluate the reasoning aspect of VQA separately from its
perception, and (2) a novel \emph{top-down calibration} technique that allows the model to answer reasoning questions
even with imperfect perception.
To this end, we introduce a \emph{differentiable first-order logic} formalism for VQA that explicitly decouples question
answering from visual perception.
On the challenging GQA dataset, this framework is used to perform in-depth, disentangled comparisons between well-known VQA models leading to informative insights regarding the participating models as well as the task. 
\end{abstract}

%% file: sections/introduction.tex
\section{Introduction}
\label{sec:intro}

\begin{figure*}[t]
    \includegraphics[width=0.85\textwidth, trim=10 100 120 200, clip]{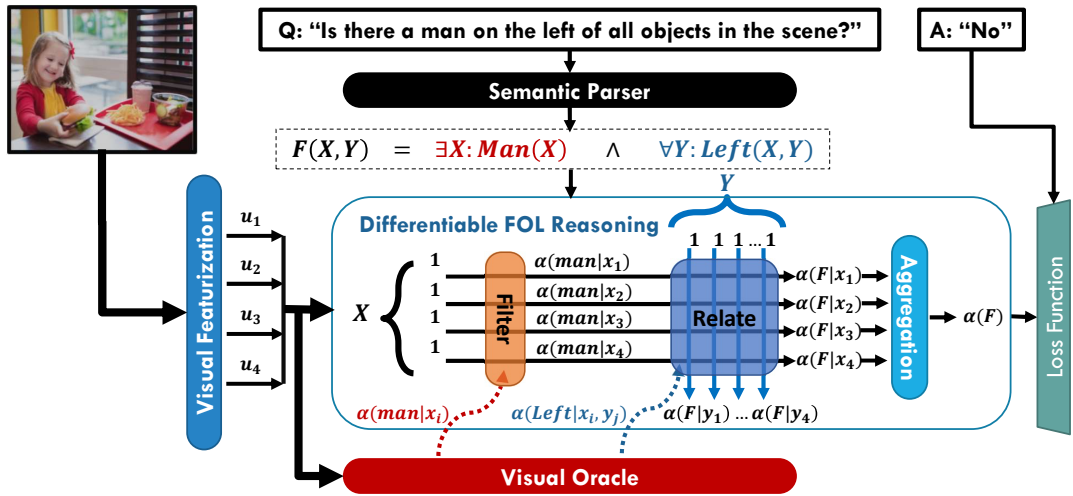}
    \centering
    \vspace{-0.5\baselineskip}
    \caption{The multi-step question answering process in the \fwname framework, based on differentiable first-order
        logic.}
    \label{fig:framework}
    \vspace{-1\baselineskip}
\end{figure*}

Visual reasoning (VR) is the ability of an autonomous system to construct a rich representation of a visual scene and
perform multi-step inference over the scene's constituents and their relationships.
It stands among the key outstanding challenges in computer vision.
Common tangible instantiations of VR include language-driven tasks such as Visual Question Answering
(VQA)~\cite{antol2015vqa} and Visual Commonsense Reasoning (VCR)~\cite{zellers2019recognition}.
Recent advances in computer vision, representation learning, and natural language processing have
enabled continued progress on VQA with a wide variety of modeling
approaches~\cite{hudson2019learning,andreas2016neural,anderson2018bottom,hudson2018compositional,tan2019lxmert}.

A defining characteristic of VR is the interaction between a \emph{perception system} (\ie object detection and scene
representation learning) and a \emph{reasoning system} (\ie question interpretation and inference grounded in the
scene).
However, this interaction is difficult to capture and assess accurately.
For example, the definition of VQA has evolved over time to eliminate language biases that impeded its robustness as a
VR metric.
The early VQA datasets were biased to real-world language priors to the extent that many questions were answerable
without looking at the image~\cite{agrawal2018don}.
Subsequent versions improved the balance but still mostly involved simple inference questions with little
requirement for multi-step reasoning.

To facilitate progress in VR, \citet{hudson2019gqa} proposed GQA, a procedurally generated VQA dataset of multi-step
inference questions.
Although GQA targets compositional multi-step reasoning, the current GQA Challenge primarily evaluates visual
perception rather than reasoning of a VQA model.
As we show in \Cref{sec:framework}, a neuro-symbolic VQA model that has access to ground-truth scene graphs achieves 96\%
accuracy on GQA.
Moreover, language interpretation (\ie semantic parsing) alone does not capture the complexity of VR due to the language
in questions being procedurally generated.
As a result, while GQA is well suited as an evaluation environment for VR (\eg for multi-modal
pretraining tasks~\cite{tan2019lxmert,zhou2019unified}), a higher GQA accuracy does not necessarily imply
a higher reasoning capability.
In this work, we supplement GQA with a \emph{differentiable first-order logic framework} \fwname that allows us to
isolate and assess the reasoning capability of a VQA model separately from its perception.

\textbf{The \fwname Framework\footnote{The PyTorch code for the \fwname framework is publicly available at \href{https://github.com/microsoft/DFOL-VQA}{https://github.com/microsoft/DFOL-VQA}.}:}
\fwname is a \emph{neuro-symbolic} VR model.
Neuro-symbolic models such as MAC~\cite{hudson2018compositional}, Neural Module Networks~\cite{andreas2016neural}, and
Neural State Machines~\cite{hudson2019learning} implement compositional multi-step inference by modeling each step as a
differentiable operator from a functional specification of the question (\ie a program) or its approximation.
This facilitates \emph{systematicity}, \emph{compositionality}, and \emph{out-of-distribution generalization} in VQA
because accurate inference of a given question commonly requires accurate inference over its constituents and entailed
questions~\cite{vedantam2019probabilistic}.
They, however, commonly operate over the latent feature representations of objects and their relations, produced by the
underlying perception module.
This entanglement not only limits interpretability of the learned neuro-symbolic inference blocks, but also limits the
reasoning techniques applicable for VQA improvement.

In contrast to SOTA neuro-symbolic approaches, \fwname \emph{fully disentangles} visual
representation learning of a VQA model from its inference mechanism, while still being end-to-end trainable with
backpropagation (see \Cref{fig:framework}).
This enables identifying GQA questions solvable via perception vs. reasoning and evaluating their
respective contributions.

\textbf{VQA Reasoning Evaluation Score:}
To assess the reasoning capability of a VQA model, we define the \emph{VQA reasoning evaluation score} as
\emph{the extent to which the model can answer a question despite imperfect visual perception}.
If the input image is noisy or the perception system is imperfect, the learned object representations do not contain
enough information to determine certain attributes of the objects.
This potentially impedes question answering and may require non-trivial reasoning.
For example, an object detection module that misclassifies wolves as huskies may impede answering
the question \textit{``Is there a husky in the living room?''}
Similarly, the question \textit{``What is behind the broken wooden chair?''} relies on the information capturing
``broken'', ``wooden'', and ``chair'' attributes in the representation of the corresponding object.
Many VQA models answer such questions nonetheless (\eg by disregarding weak attribute signals when a strong ``chair''
signal is present in a single object in the scene), which exemplifies the kind of visual reasoning we aim to assess in
VQA.
In contrast, the questions that can be answered using a pre-trained perception system and parameter-less logical
inference do not require reasoning per se as their visual representations contain all the information necessary to
answer the question.

\textbf{Contributions:}
This work makes three contributions:
\begin{itemize}[leftmargin=*, noitemsep, topsep=0pt]
    \item We introduce differentiable first-order logic as a common formalism for compositional visual reasoning and use it as a foundation for the inference in \fwname.
    \item We use \fwname to define a disentangled evaluation methodology for VQA systems to assess the informativeness of perception as well as the power of reasoning separately. To this end, we introduce a \emph{VQA reasoning evaluation score}, an augmentation of GQA evaluation
        that eliminates questions primarily resolved by perception.
        With it, we evaluate two representatives from two families of VQA models: MAC~\cite{hudson2018compositional} and LXMERT~\cite{tan2019lxmert}.
    \item As a simple way of going beyond logical reasoning, we propose \emph{top-down calibration} via the question context on the top of FOL reasoning and
        show that it improves the accuracy of \fwname on the visually hard questions.
\end{itemize}

%% file: sections/related.tex
\section{Related Work and Background}%
\label{sec:related}

\textbf{Visual Question Answering:}
VQA has been used as a front-line task to research and advance VR capabilities.
The first release of the VQA dataset~\cite{antol2015vqa} initiated annual competitions and a wide range of modeling
techniques aimed at addressing visual perception, language understanding, reasoning, and their
combination~\cite{lu2019vilbert,tan2019lxmert,hudson2018compositional,hudson2019learning,anderson2018bottom,li2019unicoder,zhou2019unified}.
To reduce the annotation effort and control the problem complexity, CLEVR~\cite{johnson2017clevr} and
GQA~\cite{hudson2019gqa} tasks propose synthetic construction of resp. scenes and questions. 

Capturing and measuring the extent of human ability in VR accurately is a significant challenge in task design as well
as modeling.
Datasets have to account for language and real-world biases, such as non-visual
and false-premise questions~\cite{ray2016question}.
VQA models, when uncontrolled, are known to ``solve'' the task by \eg exploiting language
priors~\cite{agrawal2016analyzing,zhou2015simple}.
Different techniques have been proposed to control this phenomenon.
\citet{agrawal2018don} adversarially separate the distributions of training and validation sets.
\citet{goyal2017making} balance the VQA dataset by asking human subjects to identify \emph{distractors} -- visually
similar images that yield different answers for the same questions.
Recently, \citet{selvaraju2020squinting} augment the VQA dataset with human-annotated subquestions that measure a
model's reasoning consistency in answering complex questions.
In this work, we propose another step to improve the accuracy of VQA reasoning assessment by capturing a ``hard'' subset of GQA
questions where perception produces imperfect object representations.

\textbf{Neuro-Symbolic Reasoning:}
\fwname is a \emph{neuro-symbolic reasoning model}~\cite{garcez2019neural}.
In neuro-symbolic reasoning, answer inference is defined as a chain of differentiable modules wherein
each module implements an ``operator'' from a latent functional program representation of the question.
The approach is applicable to a wide range tasks, including visual
QA~\cite{andreas2016neural,vedantam2019probabilistic,hudson2018compositional},
reading comprehension of natural language~\cite{chen2020neural}, and
querying knowledge bases, databases, or other structured sources of
information~\cite{saha-etal-2019-complex,neelakantan2015neural,neelakantan2016learning,liang2016neural}.
The operators can be learned, like in MAC~\cite{hudson2018compositional} or pre-defined, like in
NMN~\cite{andreas2016neural}.
In contrast to \emph{semantic parsing}~\cite{cacm-liang2016learning} or \emph{program
synthesis}~\cite{gulwani2017program,parisotto2016neuro}, the model does not necessarily emit a symbolic program, although it
can involve them as an intermediate step to construct the differentiable pipeline (like in \fwname).
Neuro-symbolic reasoning is also similar to \emph{neural program induction} (NPI)
\cite{reed2015neural,cai2017making,pierrot2019learning}
but the latter
requires strong supervision in the form of traces, and  the learned ``operators'' are not
always composable or interpretable.

The main benefit of neuro-symbolic models is their \emph{compositionality}.
Because the learnable parameters of individual operators are shared for all questions and subsegments of the
differentiable pipeline correspond to constituents of each question instance, the intermediate representations produced
by each module are likely composable with each other.
This, in turn, facilitates interpretability, systematicity, and out-of-distribution generalization -- commonly
challenging desiderata of reasoning systems~\cite{vedantam2019probabilistic}.
In \Cref{sec:experiments}, we demonstrate them in \fwname over VQA.

Neuro-symbolic models can be partially or fully disentangled from the representation learning of their underlying
ground-world modality (\eg vision in the case of VQA).
Partial entanglement is the most common, wherein the differentiable reasoning operates on \emph{featurizations} of the
scene objects rather than raw pixels but the featurizations are in the uninterpretable latent
space. \textit{Neural State Machine} (NSM) ~\cite{hudson2019learning} and the \textit{eXplainable and eXplicit Neural Modules} (XNM) \cite{shi2019explainable} are prominent examples of such frameworks. As for full disentanglement, there are
 \textit{Neural-Symbolic Concept Learner} (NS-CL) ~\cite{mao2019neuro} and \textit{Neural-Symbolic VQA} (NS-VQA) ~\cite{yi2018neural} which separate scene
understanding, semantic parsing, and program execution with symbolic representations in between similar to \fwname.
However, both NS-CL and NS-VQA as well as XNM are based on operators that are \textit{heuristic realization} of the \textit{task-dependent} domain specific language (DSL) of their target datasets. In contrast, we propose a \textit{task-independent, mathematical formalism} that is probabilistically derived from the first-order logic \textit{independent} of any specific DSL. 
This highlights two important differences between \fwname and NS-CL, NS-VQA, or XNM. First, compared to these frameworks, \fwname is more \textit{general-purpose}: it can implement \textit{any} DSL that is representable by FOL. Second, our proposed disentangled evaluation methodology in Section~\ref{sec:framework} requires the reasoning framework to be mathematically \textit{sound} so that we can reliably draw conclusions based off it; this is the case for our FOL inference formalism. Furthermore, while NS-CL and NS-VQA have only been evaluated on CLEVR (with synthetic scenes and a limited vocabulary), \fwname is applied to real-life scenes in GQA.

Finally, we note that outside of VR, logic-based, differentiable neuro-symbolic formalisms have been widely used to represent knowledge in neural networks ~\cite{serafini2016logic, socher2013reasoning,xu2018semantic}. A unifying framework for many of such formalisms is \textit{Differentiable Fuzzy Logics} (DFL) ~\cite{van2020analyzing} which models quantified FOL within the neural framework. Despite the similarity in formulation, the inference in DFL is generally of exponential complexity, whereas \fwname proposes a \textit{dynamic programming} strategy to perform inference in polynomial time, effectively turning it into \textit{program-based} reasoning of recent VQA frameworks. Furthermore, while these frameworks have been used to encode symbolic knowledge into the loss function, \fwname is used to specify a unique feed-forward architecture for each individual instance in the dataset; in that sense, \fwname is similar to recent neuro-symbolic frameworks proposed to tackle the SAT problem ~\cite{selsam2018learning, amizadeh2018learning, amizadeh2019pdp}.

%% file: sections/methodology.tex
\section{Differentiable First-Order Logic for VR}
\label{sec:dfol}
We begin with the formalism of differentiable first-order logic (DFOL) for VR systems, which forms the foundation for
the \fwname framework.
DFOL is a formalism for inference over statements about an image and its constituents.
It has two important properties:
(a) it \emph{disentangles} inference from perception, so that \eg the operation \textit{"filter all the red
objects in the scene"} can be split into determining the ``redness'' of every object and attending to the
ones deemed sufficiently red, and
(b) it is end-to-end differentiable, which allows training the perception system from inference results.
In \cref{sec:framework}, we show how DFOL enables us to measure reasoning capabilities of VQA models.

\subsection{Visual Perception}
Given an image $\mathcal{I}$, let $\mathcal{V}=\{\boldsymbol{v}_1,\boldsymbol{v}_2,...,\boldsymbol{v}_N\}$ be a set of
feature vectors $\boldsymbol{v}_i \in \mathbb{R}^d$ representing a set of $N$ objects detected in $\mathcal{I}$.
This detection can be done via different pre-trained models such as Faster-RCNN \cite{ren2015faster} for object
detection or Neural Motifs \cite{neuralmotifs} or Graph-RCNN \cite{yang2018graph} for scene graph generation.\footnote{$\mathcal{V}$ can also include
features of \emph{relations} between the objects. Relation features have been shown to be helpful in tasks such as image captioning and information retrieval \cite{rscan}}
As is common in VQA, we assume $\mathcal{V}$ as given, and refer to it as
the \textit{scene visual featurization}.

Furthermore, we introduce the notion of \textit{neural visual oracle} $\mathcal{O}=\{\mathcal{M}_f, \mathcal{M}_r\}$
where $\mathcal{M}_f$ and $\mathcal{M}_r$ are neural models parametrized by vectors $\boldsymbol{\theta}_f$ and
$\boldsymbol{\theta}_r$, respectively.
Conceptually, $\mathcal{M}_f(\boldsymbol{v}_i, \pi\mid \mathcal{V})$ computes the likelihood of the natural language
predicate $\pi$ holding for object $\boldsymbol{v}_i$ (e.g.
$\mathcal{M}_f(\boldsymbol{v}_i, "red"\mid \mathcal{V})$).
Similarly, $\mathcal{M}_r(\boldsymbol{v}_i, \boldsymbol{v}_j, \pi\mid \mathcal{V})$ calculates the likelihood of $\pi$
holding for a pair of objects $\boldsymbol{v}_i$ and $\boldsymbol{v}_j$ (\eg
$\mathcal{M}_r(\boldsymbol{v}_i, \boldsymbol{v}_j, "holding"\mid \mathcal{V})$).
$\mathcal{O}$ combined with the visual featurization forms the \textit{perception system} of \fwname.

\subsection{First-Order Logic over Scenes}
Given $N$ objects in the scene, we denote by the upper-case letters $X, Y, Z, ...$ categorical
variables over the objects' index set $I=\{1,...,N\}$.
The values are denoted by subscripted lower-case letters -- \eg
$X=x_i$ states that $X$ is set to refer to the $i$-th object in the scene.
The $k$-arity \textit{predicate} $\pi:I^k\mapsto \{\top, \bot\}$ defines a Boolean function on $k$
variables $X, Y, Z, ...$ defined over $I$.
In the context of visual scenes, we use unary predicates $\pi(\cdot)$ to describe object \textit{names} and
\textit{attributes} (\eg $\mathsf{Chair}(x_i)$ and $\mathsf{Red}(y_j)$), and binary predicates $\pi(\cdot,\cdot)$ to
describe \textit{relations} between pairs of objects (\eg $\mathsf{On}(y_j, x_i)$).
Given the definitions above, we naturally define \emph{quantified first-order logical (FOL) formulae} $\mathcal{F}$, \eg
\begin{equation}\label{eq:example}
    \mathcal{F}(X,Y)=\exists X,\forall Y: \mathsf{Chair}(X)\land \mathsf{Left}(X, Y)
\end{equation}
states that \textit{"There is a chair in the scene that is to the left of all other objects."}

FOL is a more \textit{compact} way to describe the visual scene than the popular \textit{scene graph}
\cite{yang2018graph} notation, which can be seen as a \textit{Propositional Logic} description of the scene, also known
as \emph{grounding} the formula.
More importantly, while scene graph is only used to \textit{describe} the scene, FOL allows us to
perform \emph{inference} over it.
For instance, the formula in Eq. \eqref{eq:example} also encodes the binary question \textit{"Is there a chair in the scene to the left of all other objects?"}
In other words, a FOL formula encodes both a \textit{descriptive statement} and a \textit{hypothetical
question} about the scene.
This is the key intuition behind \fwname and the common formalism behind its methodology.

\subsection{Inference}\label{sec:inference}
Given a NL (binary) question $\mathcal{Q}$ and a corresponding FOL formula $\mathcal{F_Q}$, the answer
$a_{\mathcal{Q}}$ is the result of evaluating $\mathcal{F_Q}$.
We reformulate this probabilistically as
\begin{equation}\label{eq:answer}
    \Pr(a_{\mathcal{Q}}=\text{``yes''}\mid \mathcal{V})=\Pr(\mathcal{F_Q}\Leftrightarrow \top\mid \mathcal{V})\triangleq
    \alpha(\mathcal{F_Q}).
\end{equation}
The na\"ive approach to calculate the probability in \cref{eq:answer} requires evaluating \textit{every}
instantiation of $\mathcal{F_Q}$, which are of exponential number.
Instead, we propose a dynamic programming strategy based on the intermediate notion of \textit{attention} which casts inference as a multi-hop execution of a
\emph{functional program} in polynomial time.

Assume $\mathcal{F_Q}$ is minimal and contains only the operators $\land$ and
$\neg$ (which are functionally complete).
We begin by defining the concept of \textit{attention} which in \fwname naturally arises by instantiating a variable in
the formula to an object:

\begin{definition}
Given a FOL formula $\mathcal{F}$ over the variables $X, Y, Z, ...$, the attention on the object $x_i$ w.r.t.
$\mathcal{F}$ is:
\begin{align}\label{eq:attention}
    \alpha(\mathcal{F}\mid x_i)&\triangleq\Pr(\mathcal{F}_{X=x_i}\Leftrightarrow \top\mid\mathcal{V}) \\
    \qquad\text{ where } \mathcal{F}_{X=x_i}&\triangleq\mathcal{F}(x_i, Y, Z,...), \forall i \in [1..N]
\end{align}
\end{definition}

Similarly, one can compute the \textit{joint attention} $\alpha(\mathcal{F}\mid x_i, y_j,...)$ by fixing more
than one variable to certain objects.
For example, given the formula in \cref{eq:example}, $\alpha(\mathcal{F}\mid x_i)$ represents the probability that
\textit{"The $i$-th object in the
scene is a chair that is to the left of all other objects."} and $\alpha(\mathcal{F}\mid y_j)$ represents the probability
that \textit{"The $j$-th object in the scene is to the right of a chair."}.

Next, we define the attention vector on variable $X$ w.r.t.
formula $\mathcal{F}$ as $\boldsymbol{\alpha}(\mathcal{F}\mid X)=[\alpha(\mathcal{F}\mid x_i)]_{i=1}^N$.
In similar way, we define the attention matrix on two variables $X$ and $Y$ w.r.t.
formula $\mathcal{F}$ as $\boldsymbol{\alpha}(\mathcal{F}\mid X,Y)=[\alpha(\mathcal{F}\mid x_i, y_j)]_{i,j=1}^N$.
Given these definitions, the following lemma gives us the first step toward calculating the likelihood in Eq.
\eqref{eq:answer} from attention values in polynomial time:

\begin{lemma}\label{lemma:ll}
Let $\mathcal{F}$ be a FOL formula with left most variable $X=LMV(\mathcal{F})$ that appears with logical quantifier
$q\in\{\exists, \forall,\nexists\}$.
Then we have:
\begin{equation}
    \alpha(\mathcal{F})=\Pr(\mathcal{F}\Leftrightarrow \top\mid \mathcal{V})=\mathcal{A}_q\big(\boldsymbol{\alpha}(\mathcal{F}\mid X)\big)
\end{equation}
where $\boldsymbol{\alpha}(\mathcal{F}\mid X)$ is the attention vector on $X$ and $\mathcal{A}_q(\cdot)$ is the
quantifier-specific aggregation function defined as:
\allowdisplaybreaks
\begin{align}
    \mathcal{A}_{\forall}(a_1, ..., a_N)&=\prod\nolimits_{i=1}^N a_i\\
    \mathcal{A}_{\exists}(a_1, ..., a_N)&=1-\prod\nolimits_{i=1}^N (1 - a_i)\\
    \mathcal{A}_{\nexists}(a_1, ..., a_N)&=\prod\nolimits_{i=1}^N (1 - a_i)
\end{align}
\end{lemma}
Furthermore, given two matrix $\boldsymbol{A}$ and $\boldsymbol{B}$, we define the \textit{matrix $Q$-product}
$\boldsymbol{C}=\boldsymbol{A}\times_{q}\boldsymbol{B}$ w.r.t.
the quantifier $q$ as:
\begin{equation}
    \boldsymbol{C}_{i,j}=[\boldsymbol{A}\times_{q}\boldsymbol{B}]_{i,j} \triangleq \mathcal{A}_q\big(A_{i\cdot}\odot B_{\cdot j}\big)
\end{equation}
where $A_{i\cdot}$ and $B_{\cdot j}$ are respectively the $i$-th row of $\boldsymbol{A}$ and the $j$-th column of $\boldsymbol{B}$, and $\odot$ denotes the Hadamard product. In general, the $Q$-product can be used to aggregate attention tensors (multi-variate logical formulas) along a certain axis (a specific variable) according to the variable's quantifier. 

Lemma \ref{lemma:ll} reduces the computation of the answer likelihood to computing the attention vector of the left most
variable w.r.t.
$\mathcal{F}$.
The latter can be further calculated recursively in polynomial time as described below.

\begin{lemma}[Base Case] \label{lemma:base}
If $\mathcal{F}$ only constitutes the literal~$\top$, the attention vector $\boldsymbol{\alpha}(\mathcal{F}\mid X)$ is the $\mathbf{1}$ vector.
\end{lemma}

\begin{lemma}[Recursion Case]\label{lemma:recursion}
We have three cases:

\textbf{(A)} \textbf{Negation Operator}:\\ if $\mathcal{F}(X, Y, Z, ...)=\neg \mathcal{G}(X, Y, Z, ...)$, then we have:
\begin{equation}
    \boldsymbol{\alpha}(\mathcal{F}\mid X)=\mathbf{1} - \boldsymbol{\alpha}(\mathcal{G}\mid X)\triangleq\mathbf{Neg}\big[\boldsymbol{\alpha}(\mathcal{G}\mid X)\big]
\end{equation}
\textbf{(B)} \textbf{Filter Operator}: if $\mathcal{F}(X, Y, Z, ...)=\pi(X)\land \mathcal{G}(X, Y, Z, ...)$ where
$\pi(\cdot)$ is a unary predicate, then:
\begin{align}\label{eq:filter}
    \boldsymbol{\alpha}(\mathcal{F}\mid X)=\boldsymbol{\alpha}(\pi\mid X)\odot \boldsymbol{\alpha}(\mathcal{G}\mid X)\triangleq\mathbf{Filter}_{\pi}\big[\boldsymbol{\alpha}(\mathcal{G}\mid X)\big]
\end{align}
\textbf{(C)} \textbf{Relate Operator}:\\ if $\mathcal{F}(X, Y, Z,
...)=\big[\bigwedge_{\pi\in\Pi_{XY}}\pi(X,Y)\big]\land\mathcal{G}(Y, Z, ...)$ where $\Pi_{XY}$ is the set of all binary
predicates defined on variables $X$ and $Y$ in $\mathcal{F}$, then we have:
\begin{align}\label{eq:relate}
    \boldsymbol{\alpha}(\mathcal{F}\mid X)&=\bigg[\bigodot_{\pi\in\Pi_{XY}}\boldsymbol{\alpha}(\pi\mid X,Y)\bigg]\times_{q}\boldsymbol{\alpha}(\mathcal{G}\mid Y)\nonumber\\&\triangleq\mathbf{Relate}_{ \Pi_{XY},q}\big[\boldsymbol{\alpha}(\mathcal{G}\mid Y)\big]
\end{align}
where $q$ is the quantifier of variable $Y$ in $\mathcal{F}$.
\end{lemma}
The attention vector $\boldsymbol{\alpha}(\pi\mid X)$ and the attention matrix $\boldsymbol{\alpha}(\pi\mid X, Y)$ in
Eqs.
\eqref{eq:filter} and \eqref{eq:relate}, respectively, form the leaves of the recursion tree and contain the
probabilities of the atomic predicate $\pi$ holding for specific object instantiations.
These probabilities are directly calculated by the visual oracle $\mathcal{O}$.
In particular, we propose:
\begin{align}
    \alpha(\pi\mid x_i)&=\mathcal{M}_f(\boldsymbol{v}_i, \pi\mid\mathcal{V}), \pi\in\Pi_u\label{eq:att}\\
    \alpha(\pi\mid x_i, y_j)&=\mathcal{M}_r(\boldsymbol{v}_i, \boldsymbol{v}_j, \pi\mid\mathcal{V}),
    \pi\in\Pi_b\label{eq:rel}
\end{align}
where $\Pi_u$ and $\Pi_b$ denote the sets of all unary and binary predicates in the model's concept dictionary.

The recursion steps in Lemma \ref{lemma:recursion} can be seen as \textit{operators} that given an
input attention vector produce an output attention vector.
In fact, \cref{eq:filter} and \cref{eq:relate} are respectively the DFOL embodiments of the abstract
\textbf{Filter} and \textbf{Relate} operations widely used in multi-hop VQA models.
In other words, by abstracting the recursion steps in Lemma \ref{lemma:recursion} into operators, we turn a
descriptive FOL formula into an executable program which can be evaluated to produce the probability
distribution of the answer.
For example, by applying the steps in Lemmas \ref{lemma:ll}-\ref{lemma:recursion} to
\cref{eq:example}, we get the following program to calculate its likelihood:
\begin{equation}
\alpha(\mathcal{F})=\mathcal{A}_{\exists}\big(\mathbf{Filter}_{\mathsf{Chair}}\big[\mathbf{Relate}_{\{\mathsf{Left}\}, \forall}[\mathbf{1}]\big]\big)
\end{equation}

\begin{algorithm}[t]
    \caption{Question answering in DFOL.}
    \label{alg:qa}
\begin{algorithmic}
    \STATE {\bfseries Input:} Question $\mathcal{F_Q}$ (binary or open), threshold $\theta$
    \IF{$\mathcal{F_Q}$ is a binary question}
        \RETURN $\alpha(\mathcal{F_Q}) > \theta$
    \ELSE
        \STATE Let $ \left\{ a_1, \dots, a_k \right\} $ be the \emph{plausible} answers for $\mathcal{F_Q}$
        \RETURN $\argmax_{1 \le i \le k} \alpha(\mathcal{F}_{\mathcal{Q},a_i})$
    \ENDIF
\end{algorithmic}
\end{algorithm}

\Cref{alg:qa} presents the final operationalization of question answering as inference over formulae in DFOL.
For open questions such as \textit{``What is the color of the chair to the left of all objects?''}, it translates them into a set of binary questions over the \emph{plausible} set of answer options (\eg all color names), which can be predefined or learned.

%% file: sections/framework.tex
\section{VQA Reasoning Evaluation Score}
\label{sec:framework}

In this section, we describe our methodology of \emph{VQA reasoning evaluation}.
Given a VQA model $\mathcal{M}$ over the visual featurization $\mathcal{V}$, our goal is to study and measure:
\begin{itemize}[nosep]
    \item [(Q1)] how informative a \textit{visual featurization} $\mathcal{V}$ is on its own to accomplish a certain
        visual reasoning task, and
    \item [(Q2)] how much the \textit{reasoning capabilities} of a model $\mathcal{M}$ can compensate for the imperfections
        in perception to accomplish a reasoning task.
\end{itemize}
To this end, we use the GQA dataset~\cite{hudson2019gqa} of multi-step functional visual questions.
The GQA dataset consists of 22M questions defined over 130K real-life images.
Each image in the Train/Validation splits is accompanied by the scene graph annotation, and each question in the
Train/Validation/Test-Dev splits comes with its equivalent program.
We translate the programs in GQA into a \emph{domain-specific language (DSL)} built on top of the four basic operators
\textbf{Filter}, \textbf{Relate}, \textbf{Neg} and $\bm{\mathcal{A}_q}$ introduced in the previous section.
The DSL covers $98\%$ of the questions in GQA.
See Appendix for its definition.

The DFOL formalism allows us to establish an \emph{upper bound on reasoning} -- the accuracy of a neuro-symbolic
VQA model when the information in its visual featurization is perfect.
To measure it, let $\mathcal{O^{*}}$ be a \emph{golden visual oracle} based on the information in the ground-truth GQA
scene graphs.
The parameter-less \fwname inference from \Cref{sec:dfol} achieves \textbf{$\bm{96\%}$ accuracy} on the GQA validation split using the golden oracle
$\mathcal{O^{*}}$ and the golden programs.
We manually inspected the remaining 4\% and found that almost all involved errors in the scene graph or the golden program.

This result not only verifies the soundness of \fwname as a probabilistic relaxation of the GQA DSL, but also
establishes that question understanding alone does not constitute the source of complexity in the compositional question
answering on GQA.
In other words, the main contributing factor to the performance of GQA models is the representation learning in their
underlying perception systems.
However, even with imperfect perception, many models successfully recover the right answer using language priors,
real-world biases, and other non-trivial learned \emph{visual reasoning}.
Using \fwname, we present a metric to quantify this phenomenon.

\textbf{Reasoning with Imperfect Perception:}\label{sec:imperfect}
Let $\mathcal{V}$ be a fixed scene featurization, often produced by \eg a pre-trained Faster-RCNN model.
Let $\mathcal{Q}$ be a GQA question and $\mathcal{F_Q}$ be its corresponding DFOL formula.
The VQA Reasoning Evaluation is based on two key observations:
\begin{enumerate}[nosep]
    \item If the probabilistic inference over $\mathcal{F_Q}$ produces a wrong answer, the
        featurization $\mathcal{V}$ does not contain enough information to correctly classify all attributes, classes,
        and relations involved in the evaluation of $\mathcal{F_Q}$.
    \item If $\mathcal{V}$ is informative enough to enable correct probabilistic inference over $\mathcal{F_Q}$, then
        $\mathcal{Q}$ is an ``easy'' question -- the right answer is accredited to perception alone.
\end{enumerate}

Let a \emph{base model} $\mathcal{M_\varnothing}$ be an evaluation of \cref{alg:qa} given some visual oracle
$\mathcal{O}$ trained and run over the features $\mathcal{V}$.
Note that the inference process of $\mathcal{M_{\varnothing}}$ described in \Cref{sec:dfol} involves \emph{no trainable
parameters}.
Thus, its accuracy stems entirely from the accuracy of $\mathcal{O}$ on the attributes/relations involved in any given
question.\footnote{This is not the same as classification accuracy of $\mathcal{O}$ in general because only a small
    fraction of objects and attributes in the scene are typically involved in any given question.}
Assuming a commonly reasonable architecture for the oracle $\mathcal{O}$ (\eg a deep feed-forward network over
$\mathcal{V}$ followed by sigmoid activation) trained end-to-end with backpropagation from the final answer through
$\mathcal{M_\varnothing}$, the accuracy of $\mathcal{M_\varnothing}$ thus indirectly captures \emph{the amount of
information in $\mathcal{V}$ directly involved in the inference of a given question} -- \ie Q1 above.

With this in mind, we arrive at the following procedure for quantifying the extent of reasoning of a VQA model
$\mathcal{M}$:
\begin{enumerate}[nosep]
    \item Fix an architecture for $\mathcal{O}$ as described above.
        We propose a standard in our experiments in \Cref{sec:experiments}.
    \item Train the oracle $\mathcal{O}$ on the Train split of GQA using backpropagation through
        $\mathcal{M_{\varnothing}}$ from the final answer.
    \item Let $T$ be a test set for GQA.
        Evaluate $\mathcal{M_{\varnothing}}$ on $T$ using the trained oracle $\mathcal{O}$ and ground-truth GQA
        programs.
    \item Let $T_e$ and $T_h$ be respectively the set of \emph{successful} and \emph{failed} questions by
        $\mathcal{M_{\varnothing}}$ (\ie $T_e \cup T_h = T$).
    \item Measure the \textbf{accuracy} of $\mathcal{M}$ on $T_h$.
    \item Measure the \textbf{error} of $\mathcal{M}$ of $T_e$.
\end{enumerate}
The \emph{easy set} $T_e$ and \emph{hard set} $T_h$ define, respectively, GQA instances where visual featurization alone
is sufficient or insufficient to arrive at the answer.
By measuring a model's \emph{accuracy on the hard set} (or \emph{error on the easy set}), we determine the extent to
which it uses the information in the featurization $\mathcal{V}$ to answer a hard question (or, resp., fails to do so on
an easily solvable question) -- \ie Q2 above.

Importantly, $\mathcal{M}$ need not be a DFOL-based model, or even a neuro-symbolic model, or even based on any notion
of a visual oracle -- we only require it to take as input the same visual features $\mathcal{V}$.
Thus, its accuracy on $T_h$ or error on $T_e$ is entirely attributable to its internal interaction between vision and
language modalities.
Furthermore, we can meaningfully compare $\mathcal{M}$'s reasoning score to that of any VQA model
$\mathcal{M'}$ that is based on the same featurization. (Although the comparison is not always ``fair'' as the
models may differ in \eg their pre-training data, it is still meaningful.)

%% file: sections/calibration.tex
\section{Top-Down Contextual Calibration}\label{sec:calibration}
\begin{figure*}[t]
    \includegraphics[width=0.8\textwidth]{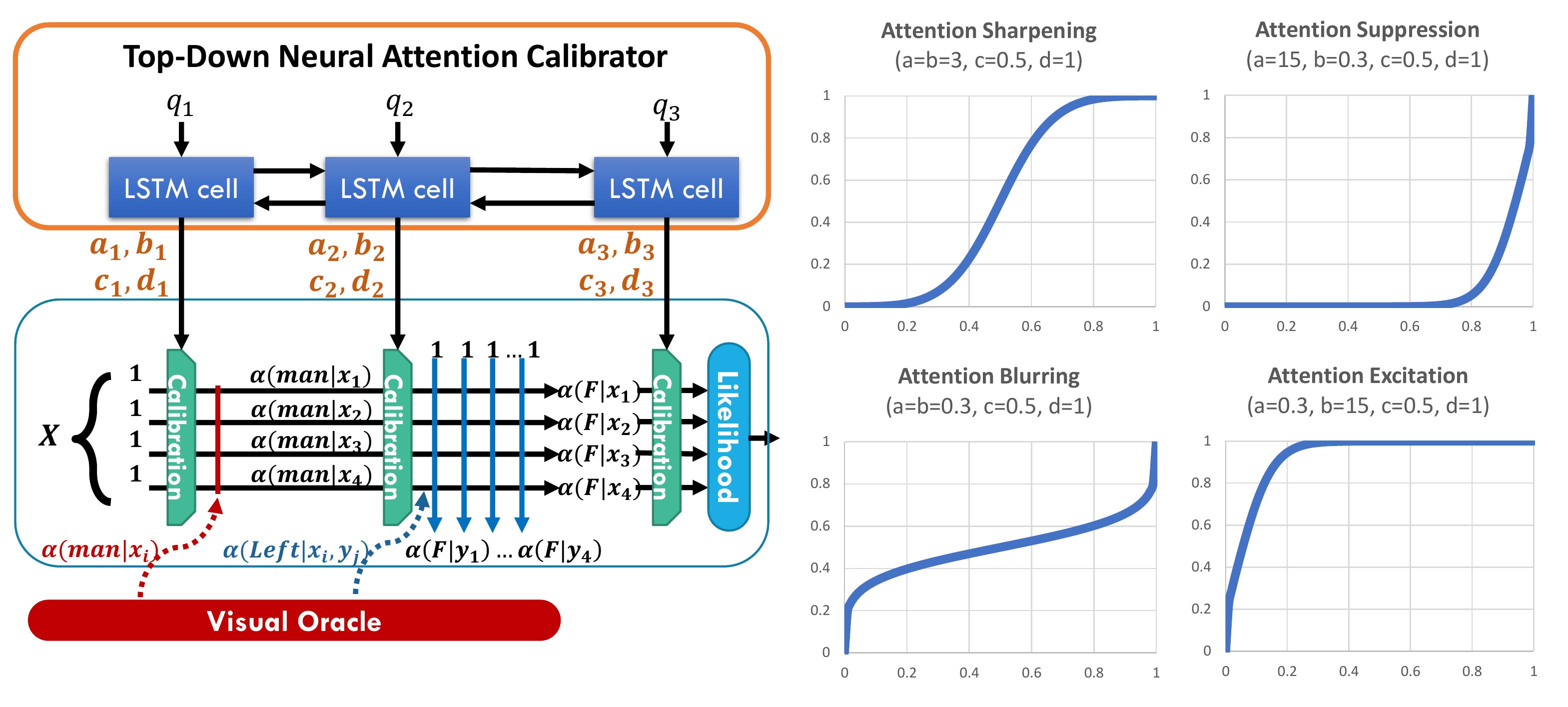}
    \centering
    \vspace{-0.5\baselineskip}
    \caption{\textbf{(Left)} The architecture of the top-down neural attention calibrator. \textbf{(Right)} Four examples of the calibration function (Eq. \eqref{eq:calibration}) shape determining to whether sharpen, blur, suppress or excite the attention values depending on the parameter values $a,b,c$ and $d$.}
    \label{fig:calibration}
    \vspace{-\baselineskip}
\end{figure*}

We now present \emph{top-down contextual calibration} as one way of \textit{augmenting} logical reasoning to compensate for imperfect perception.
Note that the FOL reasoning is a \textit{bottom-up} process in the sense that every time the oracle is queried, it does not take into consideration the broad \textit{context} of the question.
Nevertheless, considering any additional information such as the context of question can be useful especially when the
visual perception is imperfect.

Every formula $\mathcal{F}$ defines two conditional likelihoods on the attention values $\alpha(\mathcal{F}\mid x)$ over
the population of all images in the dataset: $\mathcal{P}^+_{\mathcal{F}}(\alpha) \triangleq \Pr(\alpha(\mathcal{F}\mid
x)\mid \mathcal{F}\Leftrightarrow \top)$ and $\mathcal{P}^-_{\mathcal{F}}(\alpha) \triangleq
\Pr(\alpha(\mathcal{F}\mid x)\mid \mathcal{F}\Leftrightarrow \bot)$.
In general, the bottom-up process assumes these two distributions are well separated on the extremes for every $\mathcal{F}$.
However, due to the imperfection of $\mathcal{O}$, that is not the case in practice.
The Bayesian way to address this issue is to estimate these likelihoods and use the posterior $\alpha^*(\mathcal{F}\mid x)\triangleq\Pr(\mathcal{F}\Leftrightarrow\top\mid\alpha(\mathcal{F}\mid x))$ instead of $\alpha(\mathcal{F}\mid x)$.
This is the classical notion of \textit{calibration} in binary classification \cite{platt2000probabilities}.
In our framework, we have developed the neural version of the \textit{Beta Calibration} \cite{kull2017beta} to calculate
the above posterior.
In particular, we assume the likelihoods $\mathcal{P}^+_{\mathcal{F}}(\alpha)$ and $\mathcal{P}^-_{\mathcal{F}}(\alpha)$
can be modeled as two Beta distributions with parameters $[a^+, b^+]$ and $[a^-, b^-]$, respectively.
Then, the posterior becomes $\alpha^*(\mathcal{F}\mid x)=\mathcal{C}\big(\alpha(\mathcal{F}\mid x)\big)$ where:
\begin{equation}\label{eq:calibration}
    \mathcal{C}(\alpha)=\frac{c\alpha^a}{c\alpha^a+d(1-c)(1-\alpha)^b}
\end{equation}
is called the \textit{calibration function}.
Here $a=a^+-a^-$, $b=b^--b^+$ and $c=\Pr(\mathcal{F}\Leftrightarrow\top)$ is the prior.
Furthermore, $d=B(a^+, b^+)/B(a^-, b^-)$ where $B(\cdot, \cdot)$ is the Beta function.
By $a_{\mathcal{F}}^{(i)},b_{\mathcal{F}}^{(i)},c_{\mathcal{F}}^{(i)},d_{\mathcal{F}}^{(i)}$, we denote the  parameters of the calibration function that are applied after the $i$-th operator of $\mathcal{F}$ during the attention calculation.
Instead of estimating these parameters for each possible $\mathcal{F}$ and $i$, we amortize the computation by modeling
them as a function of \textit{question context} using a Bi-LSTM \cite{huang2015bidirectional}:
\begin{equation}\label{eq:lstm}
    a_{\mathcal{F}}^{(i)},b_{\mathcal{F}}^{(i)},c_{\mathcal{F}}^{(i)},d_{\mathcal{F}}^{(i)}=\mathcal{M}_c\big(\mathcal{M}_{lstm}^{(i)}(S_{\mathcal{F}}; \boldsymbol{\theta}_{lstm}); \boldsymbol{\theta}_c\big)
\end{equation}
where $\mathcal{M}_c$ is a MLP with parameters $\boldsymbol{\theta}_c$ and $\mathcal{M}_{lstm}^{(i)}$ denotes the $i$-th
state of a Bi-LSTM parametrized by $\boldsymbol{\theta}_{lstm}$.
Here $S_{\mathcal{F}}$ denotes the context of the formula $\mathcal{F}$, which is defined as the sequence of the predicates present in the program. For example, for the formula in Eq. \eqref{eq:example}, we have $S_{\mathcal{F}}=[\mathsf{Chair},\mathsf{Left}]$. The word embedding of this context is then fed to the bi-LSTM network as its input. Figure \ref{fig:calibration} (Left) shows our proposed top-down calibration mechanism and how it affects the DFOL reasoning process. To train this calibrator, we first train the Base model \textit{without} the calibrator as before. We then freeze the weights of the visual oracle $\mathcal{O}$ in the Base model, add the calibrator on the top and run the backprop again through the resulted architecture on the training data to tune the weights of the calibrator.

Note that for parameter values $a=b=d=1$ and $c=0.5$, the calibration function in Eq. \eqref{eq:calibration} is just the Identity function; that is, the calibration function does nothing and the reasoning stays purely logical. However, as the parameters deviate from these values, so does the behavior of reasoning from the logical reasoning. Interestingly, depending on the values of its parameters, the behavior of the calibration function is quite often interpretable. In Figure \ref{fig:calibration} (Right), we have shown how the calibrator, for example, can sharpen, blur, suppress or excite visual attention values via the parameters of the calibration function. This behavior is indeed context-dependent and learned by the calibrator from data. 
For example, if the model sees the \textit{"broken wooden chair"} phrase enough times but the visual featurization is not informative enough to always detect \textit{"broken"} in the image, the calibrator may decide to \textit{excite} visual attention values upon seeing that phrase so it can make up for the imperfection of the visual system and still answer the question correctly. 

It is important to note that even though the calibrator tries to pick up informative signals from the \textit{language priors}, it does \textit{not} simply replace the visual attention values by them. Instead, it \textit{modulates} the visual attention via the language priors. So for example, if the attention values upon seeing \textit{"broken wooden chair"} is close to zero for an image (indicating that the phrase cannot be really grounded in that image), then the calibration function will not raise the attention values significantly as shown in Figure \ref{fig:calibration} (Right), even though the calibrator has learned to "excite" visual attentions for that phrase. This \textit{soft thresholding} behavior of $\mathcal{C}(\cdot)$ is entirely learned from data. Finally, we note that modulating the visual attentions via the question context is only one way of filling in the holes of perceptions. Other informative signals such as the \textit{visual context} and the \textit{feature-level, cross-modal interaction of language and vision} can be exploited to improve the accuracy of \fwname even further.

%% file: sections/experiments.tex
\section{Experiments}
\label{sec:experiments}
In this section, we experimentally demonstrate how we can incorporate our framework for evaluating the visual and the
reasoning aspects of the VQA in a decoupled manner.
To this end, we have performed experiments using our framework and candidate VQA models on the GQA dataset.

\textbf{The visual oracle:} For the experiments in this section, we have chosen a feed-forward architecture with $3$
hidden layers and an output embedding layer that covers all the concepts in the GQA vocabulary.
The weights of the embedding layer are initialized using GloVe~\cite{pennington2014glove}.

\textbf{The visual featurization:} We use the standard Faster-RCNN object featurization that is
released with the GQA dataset.
The features vectors are further augmented by the bounding box positional features for each detected object.
For binary relations, we simply concatenate the feature vectors of the two objects involved after a linear projection.
For the sake of meaningful comparison in this section, we have made sure all the participating models use the same
Faster-RCNN object featurization.

\textbf{Training setup:} For training all of \fwname models, we have used Adam optimizer with learning rate $10^{-4}$
and weight decay $10^{-10}$.
The dropout ratio is set to $0.1$.
We have also applied gradient clipping with norm $0.65$.
For better convergence, we have implemented a curriculum training scheme where we start the training with short programs
and over time we add longer programs to the training data.

\textbf{Evaluation metrics:} In addition to accuracy, we have also computed the \textit{consistency} metric as defined by the GQA Challenge ~\cite{hudson2019gqa}.

\subsection{How Informative is the GQA Visual Featurization?}
Using the settings above, we have trained the Base model $\mathcal{M}_\varnothing$.
Table \ref{tab:base} shows the accuracy and the consistency of the this model evaluated on the (balanced) Test-Dev split.
Since we wish to use the Base model to isolate only the visual informativeness of the data, we have used the golden
programs (provided in GQA) for calculating the metrics for this experiment.
Based on these results, the Faster-RCNN featurization is informative enough on its own to produce correct answers for
$51.86\%$ of the instances in the set without requiring any extra reasoning capabilities beyond FOL.
Whereas, for $48.14\%$ of the questions, the visual signal in the featurization is not informative enough to accomplish
the GQA task.
Another interesting data point here is for about $2/3$ of the binary questions, the visual features are informative
enough for question answering purposes without needing any fancy reasoning model in place, which in turn can explain why
many early classifier-based models for VQA work reasonably well on binary questions.

\subsection{Evaluating the Reasoning Capabilities of Models}
The Base model $\mathcal{M}_\varnothing$, from the previous section, can be further used to divide the test data into
the hard and easy sets as illustrated in Section \ref{sec:imperfect} (i.e.
$T_h$ and $T_e$).
In this section, we use these datasets to measure the reasoning power of candidate VQA models by calculating the metrics
$\textbf{Acc}_h$ and $\textbf{Err}_e$ as well as the consistency for each model.
See Appendix for examples of challenging instances from $T_h$ and deceptively simple instances from $T_e$.

For the comparison, we have picked two well-known representatives in the literature for which the code and
checkpoints were open-sourced.
The first is the MAC network \cite{hudson2018compositional} which belongs to the family of multi-hop,
compositional neuro-symbolic models \cite{hudson2019learning, andreas2016neural, vedantam2019probabilistic}.
The second model is the LXMERT \cite{tan2019lxmert} network which belongs to the family of Transformer-based,
vision-language models \cite{lu2019vilbert, li2019unicoder}.
Both models consume Faster-RCNN object featurization as their visual inputs and have been trained on GQA.

\begin{table}
    \centering
    \begin{tabular}{lcc}
        \hline
        Split                       & Accuracy       & Consistency      \\ \hline
        \multicolumn{1}{l|}{Open}   & 42.73 \%       & 88.74 \%         \\
        \multicolumn{1}{l|}{Binary} & 65.08 \%       & 86.65 \%         \\
        \multicolumn{1}{l|}{All}    & 51.86 \%       & 88.35 \%         \\ \hline
    \end{tabular}
    \caption{
        The Test-Dev metrics for the Base model.
        51.86\% of questions are answerable via pure FOL over Faster-RCNN features.
    }
    \vspace{-1.5\baselineskip}
    \label{tab:base}
\end{table}

\begin{table*}[t]
    \centering
    \begin{tabular}{llcccccccc}
        \hline
                        &        & \multicolumn{2}{c}{Test-Dev}          &  & \multicolumn{2}{c}{Hard Test-Dev}     &  &
                        \multicolumn{2}{c}{Easy Test-Dev}     \\ \cline{3-4} \cline{6-7} \cline{9-10}
                        & Split  & Accuracy          & Consistency       &  & $\text{Acc}_h$               & Consistency
&  & $\text{Err}_e$               & Consistency       \\ \hline
\multirow{3}{*}{MAC}    & Open   & 41.66 \%          & 82.28 \%          &  & 18.12 \%          & 74.87 \% &  &
26.70 \%          & 84.54 \% \\
                        & Binary & 71.70 \%          & 70.69 \%          &  & 58.77 \%          & 66.51 \% &  &
                        21.36 \%          & 75.37 \% \\
                        & All    & 55.37 \%          & 79.13 \%          &  & 30.54 \%          & 71.04 \% &  &
                        23.70 \%          & 82.83 \% \\ \hline
                        \multirow{3}{*}{LXMERT} & Open   & \textbf{47.02 \%} & \textbf{86.93 \%} &  & \textbf{25.27 \%} & \textbf{85.21 \%}          &  &
                        \textbf{22.92 \%} & \textbf{87.75 \%}          \\
                        & Binary & \textbf{77.63 \%} & \textbf{77.48 \%} &  & \textbf{63.02 \%} & \textbf{73.58 \%}          &  &
                        \textbf{13.93 \%} & \textbf{81.63 \%}          \\
                        & All    & \textbf{61.07 \%} & \textbf{84.48 \%} &  & \textbf{38.43 \%} & \textbf{81.05 \%}          &  &
                        \textbf{17.87 \%} & \textbf{86.52 \%}          \\ \hline
    \end{tabular}
    \caption{The test metrics for MAC and LXMERT over balanced Test-Dev and its hard
    and easy subsets according to the Base model.}
    \label{tab:mac-lxmert}
\end{table*}

\begin{table*}[t]
    \centering
    \begin{tabular}{clcccccccc}
        \hline
        \multicolumn{1}{l}{}  &        & \multicolumn{2}{c}{Test-Dev}          &           & \multicolumn{2}{c}{Hard Test-Dev}
&           & \multicolumn{2}{c}{Easy Test-Dev}    \\ \cline{3-4} \cline{6-7} \cline{9-10}
\multicolumn{1}{l}{}  & Split  & Accuracy          & Consistency       &           & $\text{Acc}_h$               &
Consistency       &           & $\text{Err}_e$              & Consistency       \\ \hline
\multirow{3}{*}{\fwname}  & Open   & \textbf{41.22 \%}          & \textbf{87.63 \%} &           & \textbf{0.53 \%}           & \textbf{11.46 \%}
&           & \textbf{2.53 \%} & \textbf{90.70 \%} \\
                      & Binary & 64.65 \%          & \textbf{85.54 \%} &           & 4.42 \%           & 61.11 \%
&           & \textbf{2.21 \%} & \textbf{86.33 \%} \\
                      & All    & 51.45 \%          & \textbf{87.22 \%} &           & 1.81 \%           & 19.44 \%
&           & \textbf{2.39 \%} & \textbf{89.90 \%} \\ \hline
\multirow{3}{*}{Calibrated \fwname} & Open   & \textbf{41.22 \%} & 86.37 \%          & \textbf{} & \textbf{0.53 \%}  &
\textbf{11.46 \%} & \textbf{} & \textbf{2.53 \%}          & 89.45 \%          \\
                      & Binary & \textbf{71.99 \%} & 79.28 \%          & \textbf{} & \textbf{37.82 \%} & \textbf{70.90
                      \%} & \textbf{} & 9.20 \%         & 84.45 \%          \\
                      & All    & \textbf{54.76 \%} & 84.48 \%          & \textbf{} & \textbf{12.91 \%} & \textbf{57.72
                      \%} & \textbf{} & 6.32 \%          & 88.51 \%          \\ \hline
  \end{tabular}
  \caption{The test metrics for \fwname and Calibrated \fwname over balanced Test-Dev
  and its hard and easy subsets.}
  \label{tab:calibration}
  \vspace{-\baselineskip}
\end{table*}

Table \ref{tab:mac-lxmert} demonstrates the various statistics obtained by evaluating the two candidate models on
balanced Test-Dev and its hard and easy subsets according to the Base model.
From these results, it is clear that LXMERT is significantly superior to MAC on the original balanced Test-Dev set.
Moreover, comparing the $\textbf{Acc}_h$ values for two models shows that the reasoning capability of LXMERT is
significantly more effective compared to that of MAC when it comes to visually vague examples.
This can be attributed to the fact that LXMERT like many other models of its family is massively pre-trained on large
volumes of vision-language bi-modal data before it is fine-tuned for the GQA task.
This pre-trained knowledge comes to the aide of the reasoning process when there are holes in the visual perception.

Another interesting observation is the comparison between the accuracy gap (i.e. $1-\textbf{Err}_e - \textbf{Acc}_h$) and the consistency gap between the hard and easy subsets for each model/split row in the table. While the accuracy gap is quite large between the two subsets (as expected), the consistency gap is much smaller (yet significant) in comparison. This shows that the notion of \textit{visual hardness (or easiness)} captured by the Base model partitioning is in fact consistent; in other words, even when VQA models struggle in the face of visually-hard examples in the hard set, their struggle is consistent across all \textit{logically-related} questions (\ie high hard consistency value in the table), which indicates that the captured notion of visual hardness is indeed meaningful. Furthermore, one may notice the smaller consistency gap of LXMERT compared to that of the MAC network, suggesting the consistent behavior of MAC is more \textit{sensitive} to the hardness level of perception compared to that of LXMERT. 

\subsection{The Effect of Top-Down Contextual Calibration}
Table \ref{tab:calibration} shows the result of applying the calibration technique from Section \ref{sec:calibration}.
Since we are using \fwname as an actual VQA model in this experiment, we have trained a simple sequence-to-sequence semantic parser to convert the natural language questions in the test set to programs.
As shown in Table \ref{tab:calibration}, the top-down calibration significantly improves the accuracy over the \fwname.
This improvement is even more significant when we look at the results on the hard set, confirming the fact that
exploiting even the simplest form of bi-modal interaction (in this case, the program context interacting with the visual
attentions) can significantly improve the performance of reasoning in the face imperfect perception.
Nevertheless, this gain comes at a cost.
Firstly, the consistency of the model over the entire set degrades.
This is, however, to be expected; after all, we are moving from pure logical reasoning to something that is not always
``logical''.
Secondly, by looking at the $\textbf{Err}_e$ values, we observe that the calibrated model starts making significant
mistakes on cases that are actually visually informative.
This reveals one of the important dangers the VQA models might fall for once they start deviating from
objective logical reasoning to attain better accuracy overall.

%% file: sections/conclusion.tex
\section{Conclusion}%
\label{sec:conclusion}

The neuro-symbolic \fwname framework, based on the differentiable first-order logic defined over the VQA task, allows us
to isolate and assess reasoning capabilities of VQA models.
Specifically, it identifies questions from the GQA dataset where the contemporary Faster-RCNN perception pipeline by
itself produces imperfect representations that do not contain enough information to answer the question via
straightforward sequential processing.
Studying these questions on the one hand motivates endeavors for improvement on the visual perception front and on the other hand provides insights into the reasoning capabilities of state-of-the-art VQA models in the face of imperfect perception as well as the sensitivity of their consistent behavior to it.
Furthermore, the accuracy and consistency on ``visually imperfect'' instances is a more accurate assessment of a model's
VR ability than dataset performance alone.
In conclusion, we believe that the methodology of vision-reasoning disentanglement, realized in \fwname, provides an
excellent tool to measure progress toward VR and some form of it should be ideally adopted by VR leaderboards.

%% file: sections/Appendix.tex
\newenvironment{changemargin}[1]{%
\begin{list}{}{%
\setlength{\topsep}{0pt}%
\setlength{\topmargin}{#1}%
\setlength{\listparindent}{\parindent}%
\setlength{\itemindent}{\parindent}%
\setlength{\parsep}{\parskip}%
}%
\item[]}{\end{list}}

\section*{Appendix A: Proofs}
\allowdisplaybreaks
\begin{proof}
\textbf{Lemma 3.1:} Let $X$ be the left most variable appearing in formula $\mathcal{F}(X, ...)$, then depending on the quantifier $q$ of $X$, we will have:
\begin{flalign*}
    \text{If } q&=\forall:
    \ \alpha(\mathcal{F})=\Pr(\mathcal{F}\Leftrightarrow \top\mid \mathcal{V})\\&=\Pr\big(\bigwedge_{i=1}^N \mathcal{F}_{X=x_i}\Leftrightarrow \top\mid\mathcal{V}\big)\\&=\prod_{i=1}^N \Pr(\mathcal{F}_{X=x_i}\Leftrightarrow \top\mid\mathcal{V})\\&=\prod_{i=1}^N \alpha(\mathcal{F}\mid x_i)=\mathcal{A}_{\forall}\big(\boldsymbol{\alpha}(\mathcal{F}\mid X)\big)&&
\end{flalign*}
\begin{flalign*}
    \text{If } q&=\exists:
    \ \alpha(\mathcal{F})=\Pr(\mathcal{F}\Leftrightarrow \top\mid \mathcal{V})\\&=\Pr\big(\bigvee_{i=1}^N \mathcal{F}_{X=x_i}\Leftrightarrow \top\mid\mathcal{V}\big)\\&=1-\prod_{i=1}^N \Pr(\mathcal{F}_{X=x_i}\Leftrightarrow \bot\mid\mathcal{V})\\&=1-\prod_{i=1}^N \big(1-\alpha(\mathcal{F}\mid x_i)\big)=\mathcal{A}_{\exists}\big(\boldsymbol{\alpha}(\mathcal{F}\mid X)\big)&&
\end{flalign*}
\begin{flalign*}
    \text{If } q&=\nexists:
    \ \alpha(\mathcal{F})=\Pr(\mathcal{F}\Leftrightarrow \top\mid \mathcal{V})\\&=\Pr\big(\bigwedge_{i=1}^N \mathcal{F}_{X=x_i}\Leftrightarrow \bot\mid\mathcal{V}\big)\\&=\prod_{i=1}^N \Pr(\mathcal{F}_{X=x_i}\Leftrightarrow \bot\mid\mathcal{V})\\&=\prod_{i=1}^N \big(1-\alpha(\mathcal{F}\mid x_i)\big)=\mathcal{A}_{\nexists}\big(\boldsymbol{\alpha}(\mathcal{F}\mid X)\big)&&
\end{flalign*}
Note that the key underlying assumption in deriving the above proofs is that the binary logical statements $\mathcal{F}_{X=x_i}$ for all objects $x_i$ are independent random variables \textit{given} the visual featurization of the scene, which is a viable assumption.
\end{proof}

\begin{proof}
\textbf{Lemma 3.2:} $\boldsymbol{\alpha}(\mathcal{F}\mid X)=\big[\Pr(\mathcal{F}_{X=x_i}\Leftrightarrow \top\mid\mathcal{V})\big]_{i=1}^N=\big[\Pr(\top\Leftrightarrow \top\mid\mathcal{V})\big]_{i=1}^N=\boldsymbol{1}$
\end{proof}

\begin{proof}
\textbf{Lemma 3.3:}
\begin{itemize}
    \item[(A)] If $\mathcal{F}(X, Y, Z, ...)=\neg \mathcal{G}(X, Y, Z, ...)$:
    \begin{flalign*}
        \boldsymbol{\alpha}(\mathcal{F}\mid X)&=\big[\Pr(\mathcal{F}_{X=x_i}\Leftrightarrow \top\mid\mathcal{V})\big]_{i=1}^N\\&=\big[\Pr(\mathcal{G}_{X=x_i}\Leftrightarrow \bot\mid\mathcal{V})\big]_{i=1}^N\\&=\big[1-\alpha(\mathcal{G}\mid x_i)\big]_{i=1}^N=\boldsymbol{1} - \boldsymbol{\alpha}(\mathcal{G}\mid X)&&
    \end{flalign*}
    \item[(B)] If $\mathcal{F}(X, Y, Z, ...)=\pi(X)\land \mathcal{G}(X, Y, Z, ...)$ where $\pi(\cdot)$ is a unary predicate:
    \begin{flalign*}
        \boldsymbol{\alpha}&(\mathcal{F}\mid X)=\big[\Pr(\mathcal{F}_{X=x_i}\Leftrightarrow \top\mid\mathcal{V})\big]_{i=1}^N\\&=\big[\Pr(\pi(x_i)\land\mathcal{G}_{X=x_i}\Leftrightarrow \top\mid\mathcal{V})\big]_{i=1}^N\\&=\big[\Pr(\pi(x_i)\Leftrightarrow \top\land\mathcal{G}_{X=x_i}\Leftrightarrow \top\mid\mathcal{V})\big]_{i=1}^N\\&=\big[\Pr(\pi(x_i)\Leftrightarrow \top\mid\mathcal{V})\cdot\Pr(\mathcal{G}_{X=x_i}\Leftrightarrow \top\mid\mathcal{V})\big]_{i=1}^N\\&=\big[\alpha(\pi\mid x_i)\cdot\alpha(\mathcal{G}\mid x_i)\big]_{i=1}^N\\&=\boldsymbol{\alpha}(\pi\mid X)\odot \boldsymbol{\alpha}(\mathcal{G}\mid X)&&
    \end{flalign*}
    \item[(C)] If $\mathcal{F}(X, Y, Z,
    ...)=\big[\bigwedge_{\pi\in\Pi_{XY}}\pi(X,Y)\big]\land\mathcal{G}(Y, Z, ...)$ where $\Pi_{XY}$ is the set of all binary predicates defined on variables $X$ and $Y$ in $\mathcal{F}$ and $Y$ is the left most variable in $\mathcal{G}$ with quantifier $q$:
    \begin{flalign*}
        &\boldsymbol{\alpha}(\mathcal{F}\mid X)=\big[\Pr(\mathcal{F}_{X=x_i}\Leftrightarrow \top\mid\mathcal{V})\big]_{i=1}^N\\&=\bigg[\Pr\bigg(\underbrace{\bigg[\bigwedge_{\pi\in\Pi_{XY}}\pi(x_i,Y)\bigg]}_{\mathcal{R}_{x_i}(Y)}\land\mathcal{G}\Leftrightarrow \top\mid\mathcal{V}\bigg)\bigg]_{i=1}^N\\&\stackrel{\text{L3.1}}{=}\big[\mathcal{A}_q\big(\boldsymbol{\alpha}(\mathcal{R}_{x_i}\land\mathcal{G}\mid Y)\big)\big]_{i=1}^N \\&\stackrel{\text{L3.3B}}{=}\big[\mathcal{A}_q\big(\boldsymbol{\alpha}(\mathcal{R}_{x_i}\mid Y)\odot\boldsymbol{\alpha}(\mathcal{G}\mid Y)\big)\big]_{i=1}^N\\&\stackrel{\text{L3.3B}}{=}\bigg[\mathcal{A}_q\bigg(\bigg[\bigodot_{\pi\in\Pi_{XY}}\boldsymbol{\alpha}(\pi_{X=x_i}\mid Y)\bigg]\odot\boldsymbol{\alpha}(\mathcal{G}\mid Y)\bigg)\bigg]_{i=1}^N\\&=\bigg[\mathcal{A}_q\bigg(\bigg[\bigodot_{\pi\in\Pi_{XY}}\boldsymbol{\alpha}(\pi\mid x_i, Y)\bigg]\odot\boldsymbol{\alpha}(\mathcal{G}\mid Y)\bigg)\bigg]_{i=1}^N\\&=\bigg[\bigodot_{\pi\in\Pi_{XY}}\boldsymbol{\alpha}(\pi\mid X,Y)\bigg]\times_{q}\boldsymbol{\alpha}(\mathcal{G}\mid Y)&&   
    \end{flalign*}
\end{itemize}
Note that the key underlying assumption in deriving the above proofs is that all the unary and binary predicates $\pi(x_i)$ and $\pi(x_i, y_i)$ for all objects $x_i$ and $y_j$ are independent binary random variables \textit{given} the visual featurization of the scene, which is a viable assumption. \qedhere
\end{proof}

\begin{sidewaystable*}[p]
\centering
\renewcommand{\arraystretch}{2}
\begin{tabularx}{\linewidth}{llll}
\toprule
\textbf{GQA OP} & \textbf{T} & \textbf{Equivalent FOL Description} & \textbf{Equivalent DFOL Program} \\ \midrule
$\textbf{GSelect}(name)[]$             & N         &$name(X)$                &$\mathbf{Filter}_{\mathsf{name}}\big[\boldsymbol{1}\big]$             \\
$\textbf{GFilter}(attr)[\boldsymbol{\alpha}_X]$             & N         &$attr(X)$                 &$\mathbf{Filter}_{\mathsf{attr}}\big[\boldsymbol{\alpha}_X\big]$             \\
$\textbf{GRelate}(name, rel)[\boldsymbol{\alpha}_X]$             & N          &$name(Y)\land rel(X, Y)$                 &$\mathbf{Filter}_{\mathsf{name}}\big[\mathbf{Relate}_{ rel,\exists}[\boldsymbol{\alpha}_X]\big]$             \\
$\textbf{GVerifyAttr}(attr)[\boldsymbol{\alpha}_X]$             & Y         &$\exists X: attr(X)$                 &$\mathcal{A}_{\exists}\big(\mathbf{Filter}_{\mathsf{attr}}[\boldsymbol{\alpha}_X]\big)$             \\
$\textbf{GVerifyRel}(name, rel)[\boldsymbol{\alpha}_X]$             & Y         & $\exists Y\exists X:name(Y)\land rel(X, Y)$                &$\mathcal{A}_{\exists}\big(\mathbf{Filter}_{\mathsf{name}}\big[\mathbf{Relate}_{ rel,\exists}[\boldsymbol{\alpha}_X]\big]\big)$             \\
$\textbf{GQuery}(category)[\boldsymbol{\alpha}_X]$             & Y          & $\big[\exists X:c(X) \text{ for } c \text{ in } category\big]$                &$\big[\mathcal{A}_{\exists}\big(\mathbf{Filter}_{\mathsf{c}}[\boldsymbol{\alpha}_X]\big) \text{ for } c \text{ in } category\big]$             \\
$\textbf{GChooseAttr}(a_1, a_2)[\boldsymbol{\alpha}_X]$             & Y         & $\big[\exists X:a(X) \text{ for } a \text{ in } [a_1,a_2]\big]$                &$\big[\mathcal{A}_{\exists}\big(\mathbf{Filter}_{\mathsf{a}}[\boldsymbol{\alpha}_X]\big) \text{ for } a \text{ in } [a_1,a_2]\big]$             \\
$\textbf{GChooseRel}(n, r_1, r_2)[\boldsymbol{\alpha}_X]$             & Y         & $\big[\exists Y\exists X:n(Y)\land r(X, Y) \text{ for } r \text{ in } [r_1,r_2]\big]$                &$\big[\mathcal{A}_{\exists}\big(\mathbf{Filter}_{\mathsf{name}}\big[\mathbf{Relate}_{ rel,\exists}[\boldsymbol{\alpha}_X]\big]\big) \text{ for } r \text{ in } [r_1,r_2]\big]$             \\
$\textbf{GExists}()[\boldsymbol{\alpha}_X]$             & Y          & $\exists X ...$                & $\mathcal{A}_{\exists}(\boldsymbol{\alpha}_X)$            \\
$\textbf{GAnd}()[\boldsymbol{\alpha}_X,\boldsymbol{\alpha}_Y]$             & Y         & $\exists X ...\land\exists Y ...$                & $\mathcal{A}_{\exists}(\boldsymbol{\alpha}_X)\cdot\mathcal{A}_{\exists}(\boldsymbol{\alpha}_Y)$             \\
$\textbf{GOr}()[\boldsymbol{\alpha}_X,\boldsymbol{\alpha}_Y]$             & Y         & $\exists X ...\lor\exists Y ...$                & $1-\big(1-\mathcal{A}_{\exists}(\boldsymbol{\alpha}_X)\big)\cdot\big(1-\mathcal{A}_{\exists}(\boldsymbol{\alpha}_Y)\big)$             \\
$\textbf{GTwoSame}(category)[\boldsymbol{\alpha}_X,\boldsymbol{\alpha}_Y]$             & Y         & $\exists X\exists Y\bigvee_{c\in category}\big(c(X)\land c(Y)\big)$                 & $\begin{aligned} &\mathcal{A}_{\exists}\big(\big[\mathcal{A}_{\exists}\big(\mathbf{Filter}_{\mathsf{c}}[\boldsymbol{\alpha}_X]\big)\cdot \mathcal{A}_{\exists}\big(\mathbf{Filter}_{\mathsf{c}}[\boldsymbol{\alpha}_Y]\big) \\ &\hspace{3.26cm} \text{ for } c \text{ in } category\big]\big)\end{aligned}$             \\
$\textbf{GTwoDifferent}(category)[\boldsymbol{\alpha}_X,\boldsymbol{\alpha}_Y]$             & Y         & $\exists X\exists Y\bigwedge_{c\in category}\big(\neg c(X)\lor \neg c(Y)\big)$                 &$1 - \textbf{GTwoSame}(category)[\boldsymbol{\alpha}_X, \boldsymbol{\alpha}_Y]$             \\
$\textbf{GAllSame}(category)[\boldsymbol{\alpha}_X]$             & Y         & $\bigvee_{c\in category}\forall X: ... \rightarrow c(X)$                 &$1 - \prod_{c\in category} \mathcal{A}_{\exists}\big(\boldsymbol{\alpha}_X \odot\mathbf{Neg}\big[\mathbf{Filter}_{\mathsf{c}}[\boldsymbol{\alpha}_X]\big]\big)$             \\ \bottomrule
\end{tabularx}
\caption{The GQA operators translated to our FOL formalism. Here the notation $\boldsymbol{\alpha}_X$ is the short form for the attention vector $\boldsymbol{\alpha}(\mathcal{F}\mid X)$ where $\mathcal{F}$ represents the formula the system has already processed up until the current operator. For the sake of simplicity, we have not included all of our GQA DSL here but the most frequent ones. Also the "Relate"-related operators are only shown for the case where the input variable $X$ is the "subject" of the relation. The formalism is the same for the "object" role case except that the order of $X$ and $Y$ are swapped in the relation. The column \textbf{T} in the table indicates whether the operator is terminal or not. The full DSL can be found at our code base: \href{https://github.com/microsoft/DFOL-VQA}{https://github.com/microsoft/DFOL-VQA}.}
\label{tab:dsl}
\end{sidewaystable*}

\section*{Appendix B: The Language System}
\begin{figure*}[t]
    \includegraphics[width=0.85\textwidth]{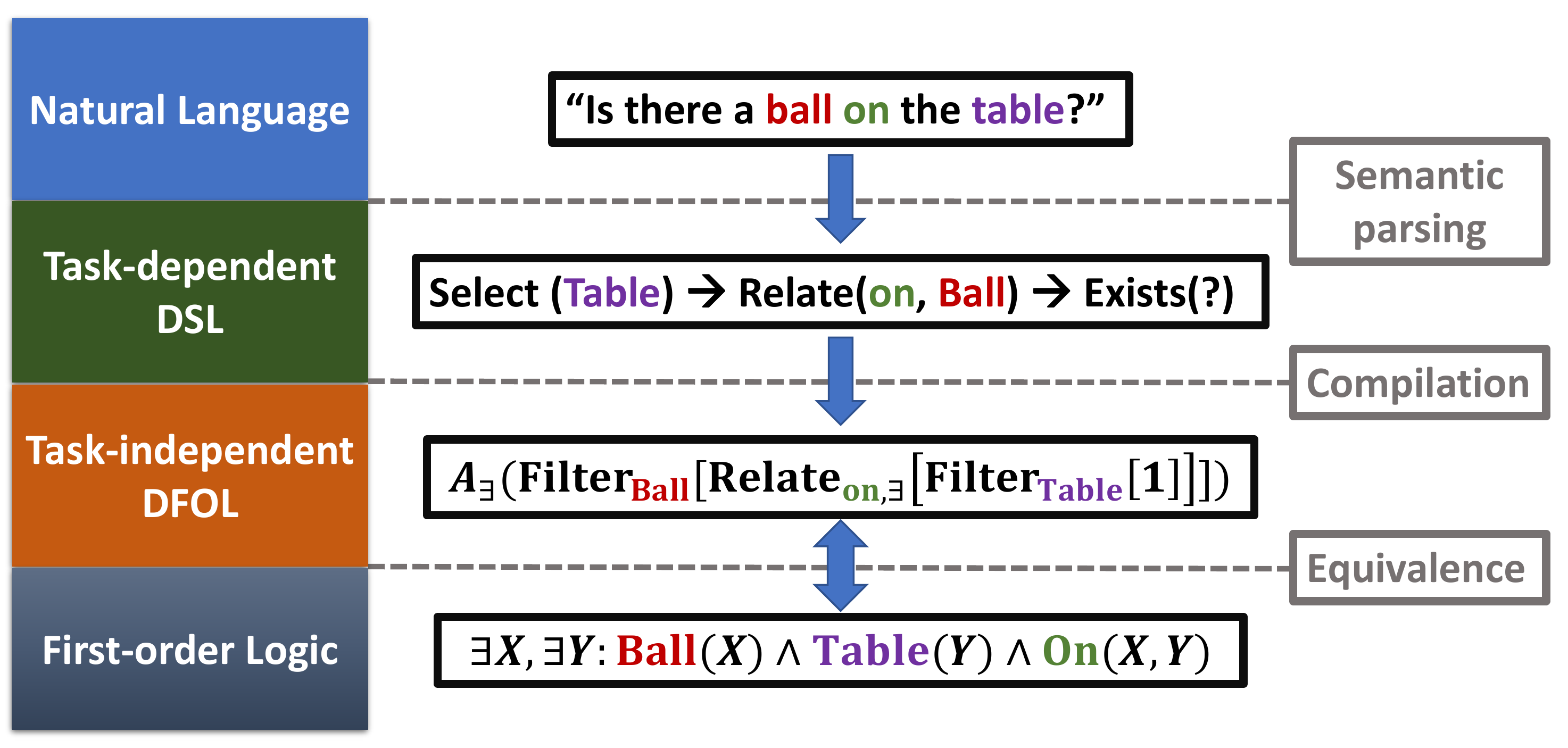}
    \centering
    \vspace{-0.5\baselineskip}
    \caption{The language system: natural language question $\xrightarrow{\text{semantic parser}}$ DSL program $\xrightarrow{\text{compiler}}$ DFOL program $\rightleftharpoons$ FOL formula.}
    \label{fig:ls}
\end{figure*}

Our language system defines the pipeline to translate the questions in the natural language (NL) all the way to the DFOL language which we can then run to find the answer to the question. However, as opposed to many similar frameworks in the literature, our translation process takes place in two steps. First, we \textit{parse} the NL question into the \textit{task-dependent}, high-level, domain-specific language (DSL) of the target task. We then \textit{compile} the resulted DSL program into the \textit{task-independent}, low-level DFOL language. This separation is important because the \fwname core reasoning engine executes the task-independent, four basic operators of the DFOL language (i.e. \textbf{Filter}, \textbf{Relate}, \textbf{Neg} and $\mathcal{A}_{\{\forall,\exists,\nexists\}}$) and \textit{not} the task specific DSL operators. This distinguishes \fwname from similar frameworks in the literature as a \textit{general-purpose} formalism; that is, \fwname can cover any reasoning task that is representable via first-order logic, and not just a specific DSL. This is mainly due to the fact that DFOL programs are equivalent to FOL formulas (up to reordering) as shown in Section 3.3. Figure \ref{fig:ls} shows the proposed language system along with its different levels of abstraction. For more details, please refer to our PyTorch code base: \href{https://github.com/microsoft/DFOL-VQA}{https://github.com/microsoft/DFOL-VQA}.   

For the GQA task, we train a neural semantic parser using the annotated programs in the dataset to accomplish the first step of translation. For the second step, we simply use a \textit{compiler}, which converts each high-level GQA operator into a composition of DFOL basic operators. Table \ref{tab:dsl} shows this (fixed) conversion along with the equivalent FOL formula for each GQA operator.

Most operators in the GQA DSL are parameterized by a set of NL tokens that specify the arguments of the operation (e.g. \textit{"attr"} in \textbf{GFilter} specifies the attribute that the operator is expected to filter the objects based upon). In addition to the NL arguments, both terminal and non-terminal operators take as input the attention vector(s) on the objects present in the scene (except for \textbf{GSelect} which does not take any input attention vector). However, in terms of their outputs, terminal and non-terminal operators are fundamentally different. A terminal operator produces a scalar likelihood or a list of scalar likelihoods (for "query" type operators). Because they are "terminal", terminal operators have logical quantifiers in their FOL description; this, in turn, prompts the aggregation operator $\mathcal{A}_{\{\forall,\exists,\nexists\}}$ in their equivalent DFOL translation. Non-terminal operators, on the other hand, produce attention vectors on the objects in the scene without calculating the aggregated likelihood.

\section*{Appendix C: Some Examples from the Hard and the Easy Sets}
\begin{figure*}[t]
\centering
    \begin{subfigure}[b]{0.475\textwidth}
        \centering
        \includegraphics[scale=0.3]{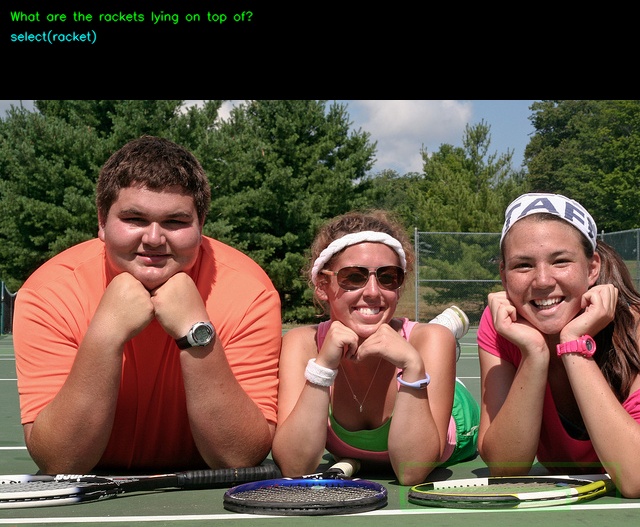}
    \caption{}
    \end{subfigure}
    \begin{subfigure}[b]{0.5\textwidth}
        \centering
        \includegraphics[scale=0.35]{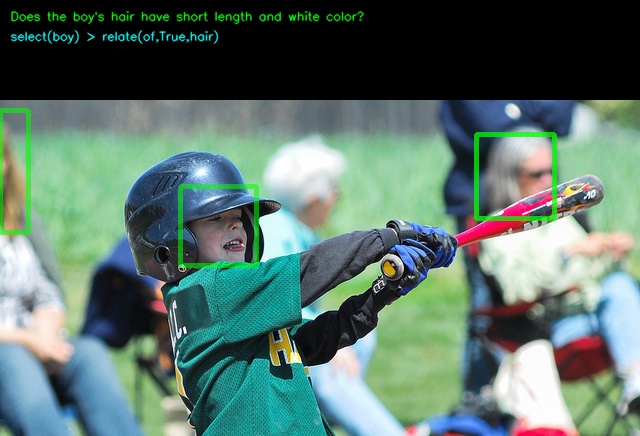}
        \caption{}
    \end{subfigure}
\caption{\textbf{Hard Set:} \textbf{(a)} Q: \textit{"What are the rackets are lying on the top of?"} As the attention bounding boxes show, the visual system has a hard time detecting the rackets in the first place and as a result is not able to reason about the rest of the question. \textbf{(b)} Q: \textit{"Does the boy's hair have short length and white color?"} In this example, the boy's hair are not even visible, so even though the model can detect the boy, it cannot detect his hair and therefore answer the question correctly.}
\label{fig:hardset1}
\end{figure*}

\begin{figure*}[t]
\centering
    \begin{subfigure}[b]{0.475\textwidth}
        \centering
        \includegraphics[scale=0.4]{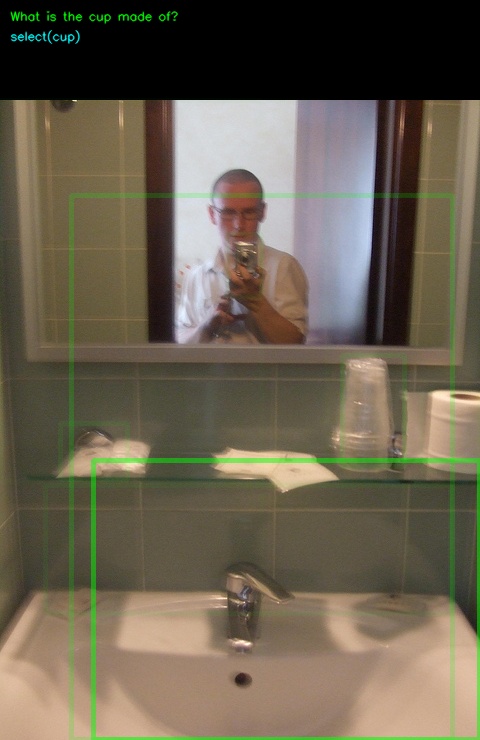}
    \caption{}
    \end{subfigure}
    \begin{subfigure}[b]{0.475\textwidth}
        \centering
        \includegraphics[scale=0.4]{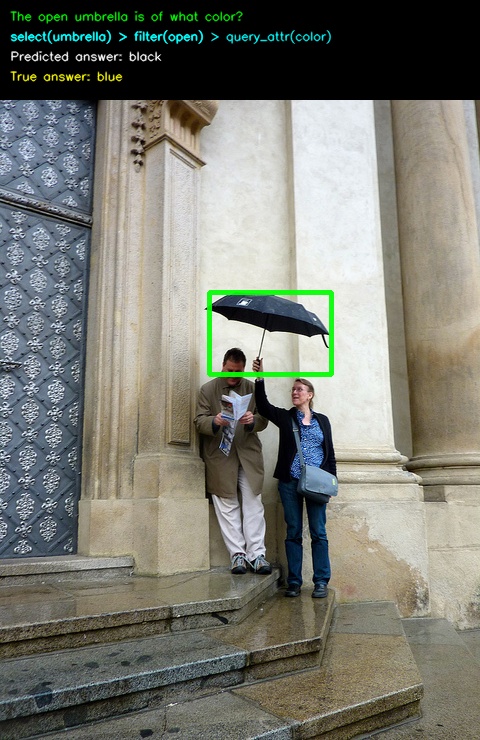}
        \caption{}
    \end{subfigure}
\caption{\textbf{Hard Set:} \textbf{(a)} Q: \textit{"What is the cup made of?"} As the attention bounding boxes show, the visual system has a hard time finding the actual cups in the first place as they are pretty blurry. \textbf{(b)} Q: \textit{"The open umbrella is of what color?"} In this example, the visual system was in fact able to detect an object that is \textit{both} "umbrella" and "open" but its color is ambiguous and can be classified as "black" even by the human eye. However, the ground truth answer is "blue" which is hard to see visually.}
\label{fig:hardset2}
\end{figure*}

\begin{figure*}[t]
\centering
    \begin{subfigure}[b]{0.475\textwidth}
        \centering
        \includegraphics[scale=0.39]{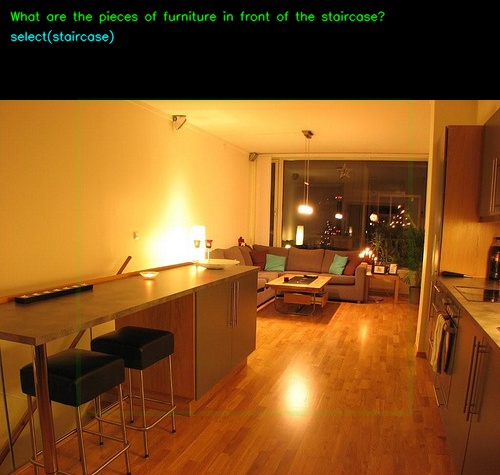}
    \caption{}
    \end{subfigure}
    \begin{subfigure}[b]{0.475\textwidth}
        \centering
        \includegraphics[scale=0.32]{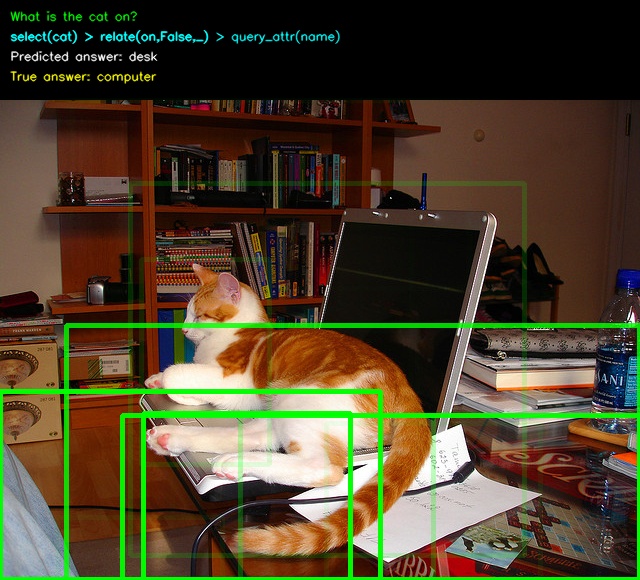}
        \caption{}
    \end{subfigure}
\caption{\textbf{Hard Set:} \textbf{(a)} Q: \textit{"What are the pieces of furniture in front of the staircase?"} In this case, the model has a hard time detecting the staircase in the scene in the first place and therefore cannot find the correct answer. \textbf{(b)} Q: \textit{"What's the cat on?"} In this example, the visual system can in fact detect the cat and supposedly the object that cat is "on"; however, it cannot infer the fact that there is actually a laptop keyboard invisible between the cat and the desk.}
\label{fig:hardset3}
\end{figure*}

\begin{figure*}[t]
\centering
    \begin{subfigure}[b]{0.475\textwidth}
        \centering
        \includegraphics[scale=0.32]{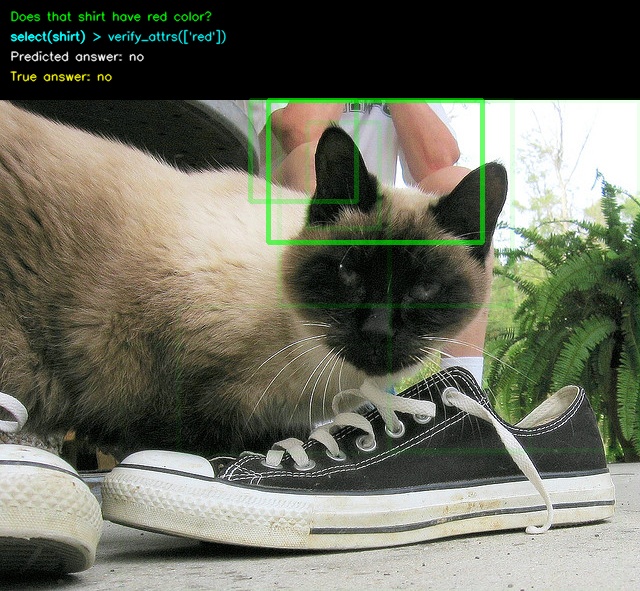}
    \caption{}
    \end{subfigure}
    \begin{subfigure}[b]{0.475\textwidth}
        \centering
        \includegraphics[scale=0.33]{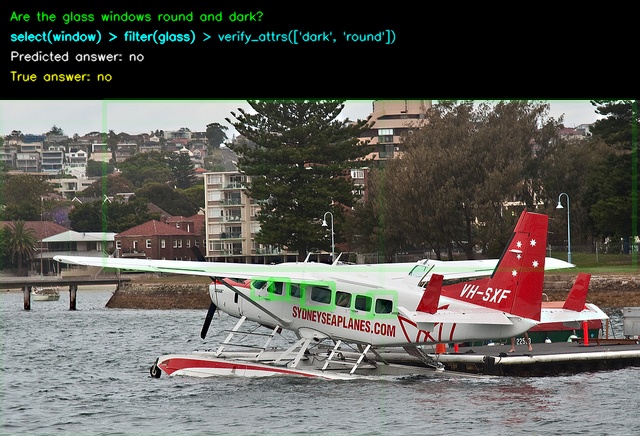}
        \caption{}
    \end{subfigure}
\caption{\textbf{Easy Set:} \textbf{(a)} Q: \textit{"Does that shirt have red color?"} \textbf{(b)} Q: \textit{"Are the glass windows round and dark?"}}
\label{fig:easyset1}
\end{figure*}

\begin{figure*}[t]
\centering
    \begin{subfigure}[b]{0.475\textwidth}
        \centering
        \includegraphics[scale=0.32]{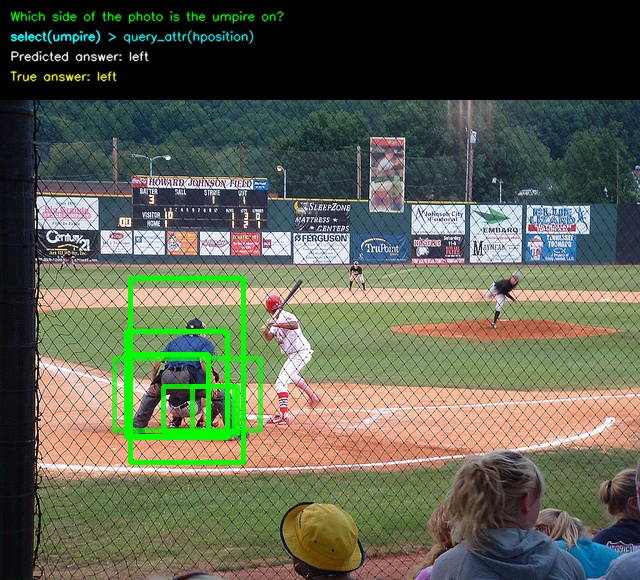}
    \caption{}
    \end{subfigure}
    \begin{subfigure}[b]{0.475\textwidth}
        \centering
        \includegraphics[scale=0.32]{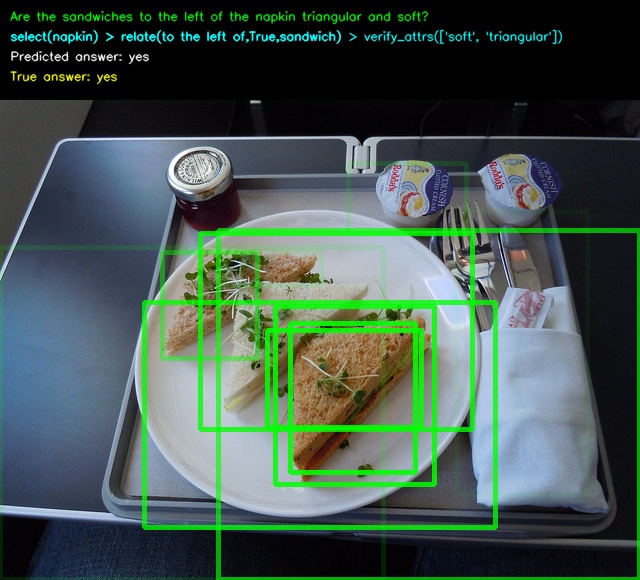}
        \caption{}
    \end{subfigure}
\caption{\textbf{Easy Set:} \textbf{(a)} Q: \textit{"What side of the photo is umpire on?"} \textbf{(b)} Q: \textit{"Are the sandwiches to the left of the napkin triangular and soft?"}}
\label{fig:easyset2}
\end{figure*}

In this appendix, we visually demonstrate a few examples from the hard and the easy subsets of the GQA Test-Dev split. Figures \ref{fig:hardset1},\ref{fig:hardset2},\ref{fig:hardset3} show a few examples from the hard set with their corresponding questions, while Figures \ref{fig:easyset1},\ref{fig:easyset2} show a few examples from the easy set. In these examples, the green rectangles represent where in the image the model is attending according to the attention vector $\boldsymbol{\alpha}(\mathcal{F}\mid X)$. Here the formula $\mathcal{F}$ represents either the entire question for the easy set examples or the partial question up until to the point where the visual system failed to produce correct likelihoods for the hard set examples. We have included the exact nature of the visual system's failure for the hard set examples in the captions. 
As illustrated in the paper, the visually hard-easy division here is with respect to the original Faster-RCNN featurization. This means that the "hard" examples presented here are \textit{not} necessarily impossible in general, but are hard with respect to this specific featurization.   

Furthermore, in Figure \ref{fig:er}, we have demonstrated two examples from the hard set for which taking into the consideration the context of the question via the calibration process helped to overcome the imperfectness of the visual system and find the correct answer. Please refer to the caption for the details. 

\begin{figure*}[t]
\begin{subfigure}[b]{0.6\textwidth}
    \centering
    \includegraphics[scale=0.9]{Figures/e1.jpg}
    \caption{}
\end{subfigure}
\begin{subfigure}[b]{0.4\textwidth}
    \centering
    \includegraphics[scale=0.29]{Figures/e3.jpg}
    \caption{}
\end{subfigure}
\caption{\textbf{(a)} Q: \textit{"Are there any lamps next to the books on the right?"} Due to the similar color of the lamp with its background, the visual oracle assigned a low probability for the predicate 'lamp' which in turn pushes the answer likelihood below $0.5$. The calibration, however, was able to correct this by considering the context of 'books' in the image. \textbf{(b)} Q: \textit{"Is the mustard on the cooked meat?"} In this case, the visual oracle had a hard time recognizing the concept of 'cooked' which in turn pushes the answer likelihood below $0.5$. The calibration, however, was able to alleviate this by considering the context of 'mustard' and 'meat' in the visual input and boosts the overall likelihood.}
\label{fig:er}
\end{figure*}